\documentclass[11pt,a4paper]{article}
\usepackage[hyperref]{acl2021}
\usepackage{times}
\usepackage{latexsym}
\usepackage{amsmath}
\usepackage{booktabs}
\usepackage{multirow}
\usepackage{graphicx}
\usepackage{amsfonts}
\usepackage{hyperref}
\usepackage{verbatim}
\usepackage{float}
\usepackage{caption}
\usepackage{subcaption}

\usepackage[ruled,vlined,linesnumbered]{algorithm2e}
\usepackage{url}
\usepackage{tabularx, multirow, multicol}
\usepackage{color}
\usepackage{soul}
\usepackage{caption}
\usepackage{float}
\usepackage{balance}

\DeclareMathOperator*{\expect}{\mathbb{E}}

\newcommand{\proposed}{OoMMix\xspace}
\newcommand{\nonlinearmix}{NonlinearMix\xspace}
\newcommand{\tmix}{TMix\xspace}
\newcommand{\mixtext}{MixText\xspace}
\newcommand{\mixuptransformer}{\textit{mixup}-transformer\xspace}

\newcommand{\tmixdag}{TMix\textsuperscript{$\dagger$}\xspace}
\newcommand{\mixtextdag}{MixText\textsuperscript{$\dagger$}\xspace}

\newcommand{\agnews}{AG News\xspace}
\newcommand{\amazon}{Amazon Review\xspace}
\newcommand{\yahoo}{Yahoo Answer\xspace}
\newcommand{\dbpedia}{DBpedia\xspace}

\newcolumntype{P}{>{\centering\arraybackslash}m{0.08\linewidth}}

\DeclareMathOperator*{\minimize}{minimize}

\usepackage{microtype}

\aclfinalcopy 
\def\aclpaperid{868} 

\title{Out-of-manifold Regularization in Contextual Embedding Space for~Text~Classification}

\author{Seonghyeon Lee\textsuperscript{1}, Dongha Lee\textsuperscript{2} and Hwanjo Yu\textsuperscript{1\thanks{ \ \ \ Corresponding author}}
 \\
  \textsuperscript{1}Dept. of Computer Science and Engineering, POSTECH, Republic of Korea \\
  \textsuperscript{2}Institute of Artificial Intelligence, POSTECH, Republic of Korea \\
  \texttt{\{sh0416,dongha.lee,hwanjoyu\}@postech.ac.kr}}

\date{}

\begin{document}
\maketitle
\begin{abstract}
Recent studies on neural networks with pre-trained weights (i.e., BERT) have mainly focused on a low-dimensional subspace, where the embedding vectors computed from input words (or their contexts) are located.
In this work, we propose a new approach to finding and regularizing the remainder of the space, referred to as out-of-manifold, which cannot be accessed through the words.
Specifically, we synthesize the out-of-manifold embeddings based on two embeddings obtained from actually-observed words, to utilize them for fine-tuning the network.
A discriminator is trained to detect whether an input embedding is located inside the manifold or not, and simultaneously, a generator is optimized to produce new embeddings that can be easily identified as out-of-manifold by the discriminator.
These two modules successfully collaborate in a unified and end-to-end manner for regularizing the out-of-manifold.
Our extensive evaluation on various text classification benchmarks demonstrates the effectiveness of our approach, as well as its good compatibility with existing data augmentation techniques which aim to enhance the manifold.

\end{abstract}

\section{Introduction}
\label{sec:intro}
Neural networks with a word embedding table have been the most popular approach to a wide range of NLP applications.
The great success of transformer-based contextual embeddings as well as masked language models~\citep{devlin-etal-2019-bert, liu2019roberta, JMLR:v21:20-074} makes it possible to exploit the pre-trained weights, fully optimized by using large-scale corpora, and it brought a major breakthrough to many problems.
For this reason, most recent work on text classification has achieved state-of-the-art performances by fine-tuning the network initialized with the pre-trained weight~\citep{devlin-etal-2019-bert}.
However, they suffer from extreme over-parameterization due to the large pre-trained weight, which allows them to be easily overfitted to its relatively small training data.

Along with outstanding performances of the pre-trained weight, researchers have tried to reveal the underlying structure encoded in its embedding space \citep{rogers2021primer}.
One of the important findings is that the contextual embeddings computed from words usually form a low-dimensional manifold \citep{ethayarajh-2019-contextual}.
In particular, a quantitative analysis on the space \citep{cai2021isotropy}, which measured the effective dimension size of BERT after applying PCA on its contextual embedding vectors, showed that 33\% of dimensions covers 80\% of the variance.
In other words, only the low-dimensional subspace is utilized for fine-tuning BERT, although a high-dimensional space (i.e., model weights with a high capacity) is provided for training.
Based on this finding on contextual embedding space, we aim to regularize the contextual embedding space for addressing the problem of over-parameterization, while focusing on the outside of the manifold (i.e., out-of-manifold) that cannot be accessed through the words.

In this work, we propose a novel approach to discovering and leveraging the out-of-manifold for contextual embedding regularization.
The key idea of our out-of-manifold regularization is to produce the embeddings that are located outside the manifold and utilize them to fine-tune the network for a target task.
To effectively interact with the contextual embedding of BERT, we adopt two additional modules, named as embedding generator and manifold discriminator.
Specifically, 1) the generator synthesizes the out-of-manifold embeddings by linearly interpolating two input embeddings computed from actually-observed words, and
2) the discriminator identifies whether an input embedding comes from the generator (i.e., the synthesized embedding) or the sequence of words (i.e., the actual embedding).
The joint optimization encourages the generator to output the out-of-manifold embeddings that can be easily distinguished from the actual embeddings by the discriminator, and the discriminator to learn the decision boundary between the in-manifold and out-of-manifold embeddings.
In the end, the fine-tuning on the synthesized out-of-manifold embeddings tightly regularizes the contextual embedding space of BERT.

The experimental results on several text classification benchmarks validate the effectiveness of our approach. 
In particular, our approach using a parameterized generator significantly outperforms the state-of-the-art mixup approach whose mixing strategy needs to be manually given by a programmer.
Furthermore, our approach shows good compatibility with various data augmentation techniques, since the target space we focus on for regularization (i.e., out-of-manifold) does not overlap with the space the data augmentation techniques have paid attention to (i.e., in-manifold).
The in-depth analyses on our modules provide an insight into how the out-of-manifold regularization manipulates the contextual embedding space of BERT.

\section{Related Work}
\label{sec:related}
In this section, we briefly review two approaches to regularizing over-parameterized network based on auxiliary tasks and auxiliary data.

\subsection{Regularization using Auxiliary Tasks}
\label{subsec:regular}
Regularization is an essential tool for good generalization capability of neural networks.
One representative regularization approach relies on designing auxiliary tasks.
\citet{liu-etal-2019-multi-task} firstly showed promising results by unifying a bunch of heterogeneous tasks and training a single unified model for all the tasks.
In particular, the synthesized task that encodes desirable features or removes undesirable features turns out to be helpful for network regularization.
\citet{devlin-etal-2019-bert} introduced the task which restores masked sentences, termed as masked language model, to encode the distributional semantic in the network;
this considerably boosts the overall performance of NLP applications.
In addition, \citet{Clark2020ELECTRA:} regularized the network by discriminating generated tokens from a language model, and \citet{NEURIPS2018_e555ebe0} utilized an additional discriminator to remove the information about word frequency implicitly encoded in the word embeddings.

\subsection{Regularization using Auxiliary Data}
\label{subsec:dataaug}
Another approach to network regularization is to take advantage of auxiliary data, mainly obtained by data augmentation, which eventually supplements the input data space.
Inspired by \citep{pmlr-v15-bengio11b} that additionally trained the network with noised (i.e., augmented) images in computer vision,
\citet{wei-zou-2019-eda} simply augmented sentences by adding a small perturbation to the original sentences, such as adding, deleting, and swapping words within the sentences.
Recent work tried to further exploit the knowledge from a pre-trained model for augmenting the sentences:
sentence back translation by using a pre-trained translation model~\citep{xie2019unsupervised}, and
masked sentence reconstruction by using a pre-trained masked language model~\citep{ng-etal-2020-ssmba}.
 
Mixup~\citep{zhang2018mixup} is also a kind of data augmentation but differs in that it performs linear interpolation on multiple input sentences and their corresponding labels.
\citet{pmlr-v97-verma19a} validated that mixup in the hidden space (instead of the input space) is also effective for regularization, and
\citet{guo2019mixup} found that mixup of images can regularize the out-of-manifold in image representations.
In the case of NLP domain, \citet{guo2019augmenting} and \citet{guo2020nonlinear} firstly adopted mixup to text data for text classification, using the traditional networks such as CNN and LSTM;
they sample their mixing coefficients from the beta distribution at the sentence-level and at the word-level, respectively.
To fully utilize the contextual embedding of transformer-based networks, \citet{chen-etal-2020-mixtext} applied mixup in the word-level contextual embedding space using a pre-trained language model (i.e., BERT), whereas
\citet{sun-etal-2020-mixup} focused on mixup in the sentence-level embedding space specifically for improving GLUE score.

\section{Method}
\label{sec:method}
\begin{figure*}
\centering
\includegraphics[width=0.99\textwidth]{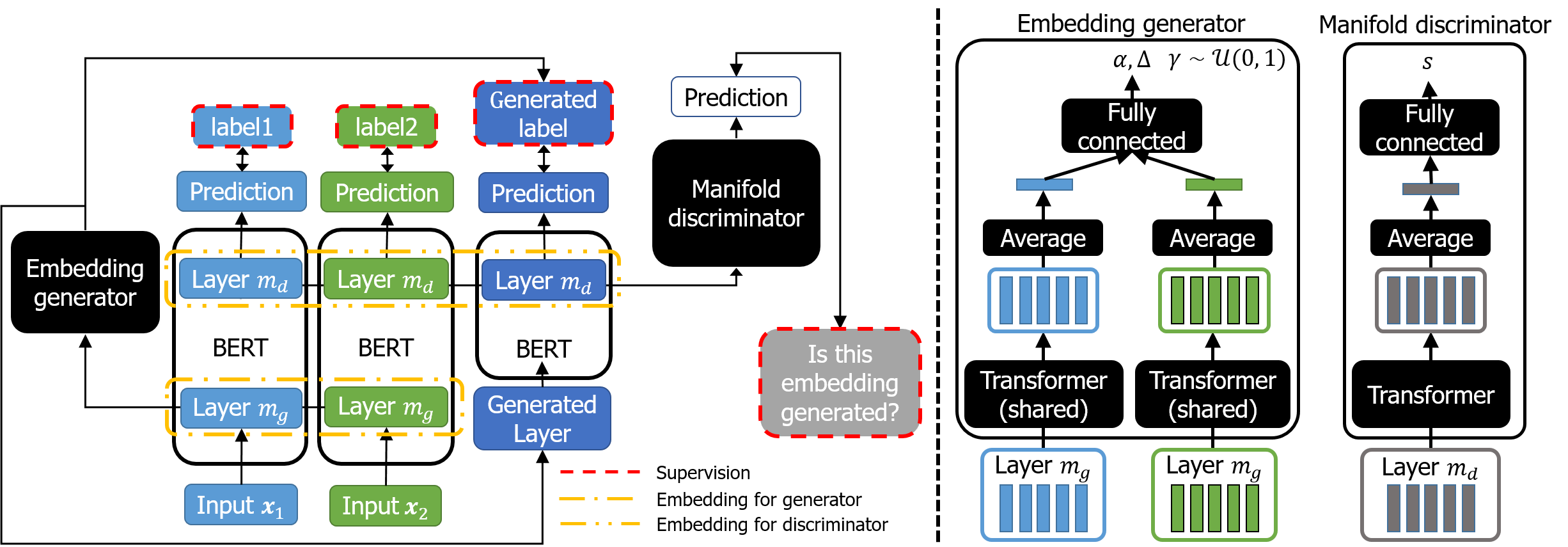}
\caption{The overview of \proposed for fine-tuning BERT (Left) and the structure of our embedding generator and manifold discriminator (Right).
}
\end{figure*}

In this section, we propose a novel mixup approach, termed as \proposed, to regularize the out-of-manifold in contextual embedding space for text classification.
We first briefly remind the architecture of BERT, then introduce two modules used for out-of-manifold regularization, which are embedding generator and manifold discriminator.

\subsection{Preliminary}
\label{subsec:preliminary}
BERT is a stack of $M$ transformer encoders pre-trained on the objective of the masked language model~\citep{devlin-etal-2019-bert}.
First, a raw sentence is split into the sequence of tokens $\mathbf{x} \in \left\{ 0, ..., \left|V\right| \right\}^L$ using a tokenizer with the vocabulary $V$, where $L$ is the sequence length.
Each token is mapped into a $D$-dimensional vector based on the embedding table.
The sequence of embedding vectors $\mathbf{h}^{(0)} \in \mathbb{R}^{L \times D}$ is transformed into the $m$-th contextual embedding $\mathbf{h}^{(m)} \in \mathbb{R}^{L \times D}$ by $m$ transformer layers~\citep{NIPS2017_3f5ee243}.

We fine-tune the pre-trained weight to classify input texts into $C$ classes.
A classifier produces the classification probability vector $o \in \mathbb{R}^{C}$ using the last contextual embedding $\mathbf{h}^{(M)}$.
Then, the optimization problem is defined based on a labeled dataset $D = \left\{ \left( \mathbf{x}_1, y_1 \right), ..., \left( \mathbf{x}_N, y_N \right) \right\}$.
\begin{gather}
    \minimize_{w_f} \underset{\left( \mathbf{x}, y \right) \in D}{\mathbb{E}} \Big[ \mathcal{L}_{C} \left( \mathbf{x}, y \right) \Big] \nonumber \\
    \mathcal{L}_{C} \left( \mathbf{x}, y \right) := \mathcal{L}_{kl} \left( f \left( \mathbf{x} \right), \mathbf{e}_y \right) \nonumber
\end{gather}
where $\mathcal{L}_{kl}$ is the Kullback-Leibler divergence and $\mathbf{e}_y \in \mathbb{R}^{C}$ is a one-hot vector representing the label $y$. 
The function $f$ is the whole process from $\mathbf{h}^{(0)}$ to $o$, called a target model, and $w_f$ is the trainable parameters for the function $f$, including the pre-trained weight of BERT and the parameters in the classifier.
For notation, $f$ can be split into several sub-processes $f(\mathbf{x}) = ( f_{m'} \circ h_m^{m'} \circ h_0^{m})(\mathbf{x})$ where $h_m^{m'} ( \mathbf{x} )$ maps the $m$-th contextual embedding into the $m'$-th contextual embedding through the layers.

\subsection{Embedding Generator}
\label{subsec:meg}
The goal of our generator network $G$ is to synthesize an artificial contextual embedding by taking two contextual embeddings (obtained from layer $m_g$) as its input.
We use linear interpolation so that the new embedding belongs to the line segment defined by the two input embeddings.
Since we limit the search space, the generator produces a single scalar value $\lambda \in \left[ 0, 1 \right]$, called a mixing coefficient.
\begin{equation*}
\begin{split}
    G \left( \mathbf{h}_1^{(m_g)}, \mathbf{h}_2^{(m_g)} \right) &= \lambda \cdot \mathbf{h}_1^{(m_g)} + \left( 1 - \lambda \right)\cdot \mathbf{h}_2^{(m_g)}\\
    \lambda &= g \left( \mathbf{h}_1^{(m_g)}, \mathbf{h}_2^{(m_g)} \right)
\end{split}
\end{equation*}
We introduce the distribution of the mixing coefficient to model its uncertainty.
To this end, our generator network produces the lower bound $\alpha$ and the interval $\Delta$ by using $\mathbf{h}_1^{(m_g)}$ and $\mathbf{h}_2^{(m_g)}$, so as to sample the mixing coefficient from the uniform distribution $\mathcal{U} \left( \alpha, \alpha + \Delta \right)$.

To avoid massive computational overhead incurred by the concatenation of two input sequences~\citep{reimers-gurevych-2019-sentence}, we adopt the Siamese architecture that uses the shared weights on two different inputs.
The generator first transforms each sequence of contextual embedding vectors by using a single transformer layer, then obtains the sentence-level embedding by averaging all the embedding vectors in the sequence.
From the two sentence-level embeddings $\mathbf{s}_1, \mathbf{s}_2 \in \mathbb{R}^{D}$, the generator obtains the concatenated embedding $\mathbf{s} = \mathbf{s}_1 \oplus \mathbf{s}_2 \in \mathbb{R}^{2D}$ and calculates $\alpha$ and $\Delta$ by using a two-layer fully-connected network with the softmax normalization.
Specifically, the last fully-connected layer outputs a normalized 3-dimensional vector, whose first and second values become $\alpha$ and $\Delta$, thereby the range of sampling distribution $(\alpha, \alpha + \Delta)$ lies in $\left[ 0, 1 \right]$.
In this work, we consider the structure of the generator to efficiently process the sequential input, but any other structures focusing on different aspects (e.g. the network that enlarges the search space) can be used as well.
For effective optimization of $\lambda$ sampled from $\mathcal{U} \left( \alpha, \alpha + \Delta \right)$, we apply the re-parameterization trick which decouples the sampling process from the computational graph \citep{Kingma2014}.
That is, we compute the mixing coefficient by using $\gamma\sim\mathcal{U}\left( 0, 1 \right)$.
\begin{align*}
    \lambda = \alpha + \gamma \times \Delta
\end{align*}

The optimization problem for text classification can be extended to the new embeddings and their labels, provided by the generator network.
\begin{gather}
    \minimize_{w_{f_{m_g}}, w_g} \underset{\left( \mathbf{x}_1, y_1 \right) \in D}{\mathbb{E}} \Big[ \mathcal{L}_{G} \left( \mathbf{x}_1, y_1 \right) \Big] \label{eq:target_obj} \\
    \mathcal{L}_{G} \left( \mathbf{x}_1, y_1 \right) := \expect_{\left( \mathbf{x}_2, y_2 \right) \in \mathcal{D}} \left[ \mathcal{L}_{kl} ( f_{m_g} ( \tilde{\mathbf{h}} ), \tilde{\mathbf{y}} ) \right] \nonumber
\end{gather}
\begin{equation*}
\begin{split}
    \lambda &\sim g\left( h_0^{m_g} \left( \mathbf{x}_1 \right), h_0^{m_g} \left( \mathbf{x}_2 \right)\right) \\
    \tilde{\mathbf{h}} &:= \lambda \cdot  h_0^{m_g} \left( \mathbf{x}_1 \right) + (1-\lambda) \cdot h_0^{m_g} \left( \mathbf{x}_2 \right)  \nonumber \\
    \tilde{\mathbf{y}} &:= \lambda \cdot \mathbf{e}_{y_1} + \left( 1 - \lambda \right) \cdot \mathbf{e}_{y_2}
\end{split}
\end{equation*}
where $w_{f_{m_g}}$ is the trainable parameters of the function $f_{m_g}$ (i.e., the process from $\mathbf{h}^{(m_g)}$ to $o$), and $w_G$ is the ones for the generator.
Similar to other mixup techniques, we impose the mixed label on the generated embedding.

\subsection{Manifold Discriminator}
\label{subsec:oomdis}
We found that the supervision from the objective~\eqref{eq:target_obj} is not enough to train the generator.
The objective optimizes the generator to produce the embeddings that are helpful for the target classification.
However, since the over-parameterized network tends to memorize all training data, the target model also simply memorizes the original data to minimize Equation~\eqref{eq:target_obj}.
In this situation, the generator is more likely to mimic the embeddings seen in the training set (memorized by the target model) rather than generate novel embeddings.
For this reason, we need more useful supervision for the generator, to make it output the out-of-manifold embeddings.

To tackle this challenge, we define an additional task that identifies whether a contextual embedding comes from the generator or actual words.
The purpose of this task is to learn the discriminative features between actual embeddings and generated embeddings, in order that we can easily discover the subspace which cannot be accessed through the actually-observed words.
For this task, we introduce a discriminator network $D$ that serves as a binary classifier in the contextual embedding space of the $m_d$-th transformer layer.

The discriminator takes a contextual embedding $\mathbf{h}^{(m_d)}$ and calculates the score $s \in \left[ 0, 1 \right]$ which indicates the probability that $\mathbf{h}^{(m_d)}$ comes from an actual sentence (i.e., $\mathbf{h}^{(m_d)}$ is located inside the manifold).
Its network structure is similar to that of the generator, except that the concatenation is not needed and the output of the two-layer fully connected network produces a single scalar value.
As discussed in Section~\ref{subsec:meg}, any network structures for focusing on different aspects can be employed.

The optimization of the generator and discriminator for this task is described as follows.
\begin{gather}
    \minimize_{w_g, w_d} \underset{\left( \mathbf{x}_1, y_1 \right) \in D}{\mathbb{E}} \Big[ \mathcal{L}_D \left( \mathbf{x}_1 \right) \Big] \label{eq:oom_obj}\\
    \mathcal{L}_D \left( \mathbf{x}_1 \right) := \underset{\left( \mathbf{x}_2, y_2 \right) \in D}{\mathbb{E}} \Big[ \mathcal{L}_{bce} ( D ( h_{m_g}^{m_d} ( \tilde{\mathbf{h}} ) ) , 0 )  \nonumber\\
    \qquad\qquad\qquad\qquad + \mathcal{L}_{bce} \left(  D \left( h_{0}^{m_d} \left( \mathbf{x} \right) \right) , 1 \right) \Big] \nonumber
\end{gather}
where $\mathcal{L}_{bce}$ is the binary cross entropy loss.
By minimizing this objective, our generator can produce the out-of-manifold embeddings that are clearly distinguished from the actual (in-manifold) contextual embeddings by the discriminator.

\subsection{Training}
\label{subsec:train}
We jointly optimize the two objectives to train the embedding generator.
Equation~\eqref{eq:target_obj} encourages the generator to produce the embeddings which are helpful for the target task, while Equation~\eqref{eq:oom_obj} makes the generator produce the new embeddings different from the contextual embeddings obtained from the words.
The final objective is defined by
\begin{equation}
    \underset{\left( \mathbf{x}, y \right) \sim D}{\mathbb{E}} \left[ \mathcal{L}_C \left( \mathbf{x}, y \right) + \mathcal{L}_G \left( \mathbf{x}, y \right) + e \mathcal{L}_D ( \mathbf{x} ) \right] \nonumber
\end{equation}
where $e$ regulates the two objectives.
The generator and discriminator collaboratively search out informative out-of-manifold embeddings for the target task while being optimized with the target model, thereby the generated embeddings can effectively regularize the out-of-manifold.

\section{Experiments}
\label{sec:exp}
\begin{table*}
\centering
\resizebox{0.75\linewidth}{!}{%
\begin{tabular}{@{}ccccccc@{}}
\toprule
Dataset & Input sentence & Class & Valid size & Valid length & Test size & Test length \\ \midrule
\agnews & content & 4 & 7.6K & 43.49 & 7.6K & 43.21 \\
\amazon & review text & 2 & 8K & 95.94 & 400K & 95.62 \\
\yahoo & title, question, answer & 10 & 50K & 109.81 & 60K & 110.74 \\
\dbpedia & content & 14 & 28K & 63.62 & 70K & 63.61 \\ \bottomrule
\end{tabular}
}
\caption{Statistics of datasets}
\label{tab:dataset}
\end{table*}
\begin{table*}
\centering
\resizebox{0.99\linewidth}{!}{%
\begin{tabular}{@{}ccccccccc@{}}
\toprule
Dataset & Train & Original & \nonlinearmix & \mixuptransformer & \tmix & \proposed$\textsuperscript{\ \ \ \ }$ & $\text{\tmix}^{\dagger}$ & $\text{\mixtext}^{\dagger}$ \\ \midrule
\multirow{3}{*}{\agnews} & 0.5K & $88.22\pm0.02$ & $88.24\pm0.05$ & $\mathbf{88.58\pm0.02}$ & $88.45\pm0.02$ & $88.41\pm0.05\textsuperscript{\ \ \ \ }$ & - & - \\
 & 2.5K & $89.92\pm0.15$ & $88.75\pm0.36$ & $89.62\pm0.09$ & $90.07\pm0.09$ & $\mathbf{90.25\pm0.05\textsuperscript{*\ \ }}$ & - & - \\
 & 10K & $91.50\pm0.05$ & $88.86\pm0.12$ & $91.37\pm0.21$ & $91.51\pm0.08$ & $\mathbf{91.83\pm0.09\textsuperscript{**}}$ & $91.0$ & $91.5$ (+20K) \\ \midrule
\multirow{3}{*}{\amazon} & 0.5K & $89.17\pm0.35$ & $89.02\pm0.21$ & $89.31\pm0.14$ & $89.57\pm0.02$ & $\mathbf{89.66\pm0.01\textsuperscript{\ \ \ \ }}$ & - & - \\
 & 2.5K & $90.96\pm0.05$ & $91.04\pm0.11$ & $90.70\pm0.05$ & $91.24\pm0.13$ & $\mathbf{91.28\pm0.12\textsuperscript{\ \ \ \ }}$ & - & - \\
 & 10K & $92.81\pm0.05$ & $91.15\pm0.42$ & $92.12\pm0.28$ & $92.79\pm0.07$ & $\mathbf{92.94\pm0.06\textsuperscript{\ \ \ \ }}$ & - & - \\ \midrule
\multirow{3}{*}{\yahoo} & 0.5K & $67.24\pm0.07$ & $67.56\pm0.37$ & $67.62\pm0.06$ & $67.57\pm0.11$ & $\mathbf{67.95\pm0.16\textsuperscript{\ \ \ \ }}$ & - & - \\
 & 2K & $70.41\pm0.04$ & $69.17\pm0.11$ & $70.29\pm0.14$ & $70.68\pm0.15$ & $\mathbf{71.08\pm0.10\textsuperscript{*\ \ }}$ & $69.8$ & $71.3$ (+50K) \\
 & 25K & $73.68\pm0.03$ & $69.31\pm0.37$ & $73.52\pm0.05$ & $73.84\pm0.00$ & $\mathbf{74.13\pm0.06\textsuperscript{*\ \ }}$ & $73.5$ & $74.1$ (+50K) \\ \midrule
\multirow{3}{*}{\dbpedia} & 0.5K & $97.86\pm0.07$ & $97.50\pm0.25$ & $98.06\pm0.05$ & $98.15\pm0.10$ & $\mathbf{98.26\pm0.04\textsuperscript{\ \ \ \ }}$ & - & - \\
 & 2.8K & $\mathbf{98.83\pm0.03}$ & $98.74\pm0.09$ & $98.76\pm0.01$ & $98.82\pm0.04$ & $\mathbf{98.83\pm0.05\textsuperscript{\ \ \ \ }}$ & $98.7$ & $98.9$ (+70K) \\
 & 35K & $98.96\pm0.07$ & $98.89\pm0.01$ & $98.91\pm0.03$ & $98.97\pm0.03$ & $\mathbf{99.03\pm0.03\textsuperscript{*\ \ }}$ & $99.0$ & $99.2$ (+70K) \\ \bottomrule
\end{tabular}%
}
\caption{Classification accuracy on sentence classification benchmarks. 
* and ** respectively indicate $p\leq 0.05$ and $p\leq 0.01$ for the paired t-test of \proposed vs. the best competitor.
\tmixdag and \mixtextdag report the scores presented in \citep{chen-etal-2020-mixtext}, where the sizes of domain-related unlabeled data are described in the parenthesis.}
\label{tab:overall}
\end{table*}
In this section, we present the experimental results supporting the superiority of \proposed among the recent mixup approaches in text classification.
Also, we investigate its compatibility with other data augmentation techniques. 
Finally, we provide in-depth analyses on our approach to further validate the effect of out-of-manifold regularization. 

\subsection{Experimental setup}
Our experiments consider 4 sentence classification benchmarks~\citep{NIPS2015_250cf8b5} of various scales.
The statistics of the datasets are summarized in Table \ref{tab:dataset}.
We follow the experimental setup used in~\citep{chen-etal-2020-mixtext} to directly compare the results with ours.
Specifically, we split the whole training set into training/validation sets, while leaving out the official test set for evaluation.
We choose the classification accuracy as the evaluation metric, considering the datasets are already class-balanced.
For the various sizes of training set from 0.5K to 35K, we apply stratified sampling to preserve the balanced class distributions.

In terms of optimization, we use BERT provided by huggingface for the classification tasks.\footnote{In our experiments, we use the checkpoint \texttt{bert-base-} \texttt{uncased} as the pre-trained weight.}
The Adam optimizer is used to fine-tune BERT with the linear warm-up for the first 1000 iterations, and the initial learning rates for the pre-trained weight and the target classifier are set to 2e-5 and 1e-3, respectively.
We set the batch size to 12 and the dropout probability to 0.1.
We attach the generator and discriminator at the third layer ($m_g=3$) and the last layer ($m_d=12$), respectively.
The two objectives equally contribute to training the generator, $e=1$, but we increase the $e$ value if the discriminator fails to discriminate the embeddings.
The accuracy is evaluated on validation set every 200 iterations, and stop training when the accuracy does not increase for 10 consecutive evaluations.
We report the classification accuracy on the test set at the best validation checkpoint and repeat the experiment three times with different random seeds to report the average with its standard deviation.
We implement the code using PyTorch and use NVIDIA Titan Xp for parallel computation.
In our environment, the training spends about 30 minutes to 3 hours depending on the dataset. 

\begin{figure*}[ht]
    \centering
    \includegraphics[width=\textwidth]{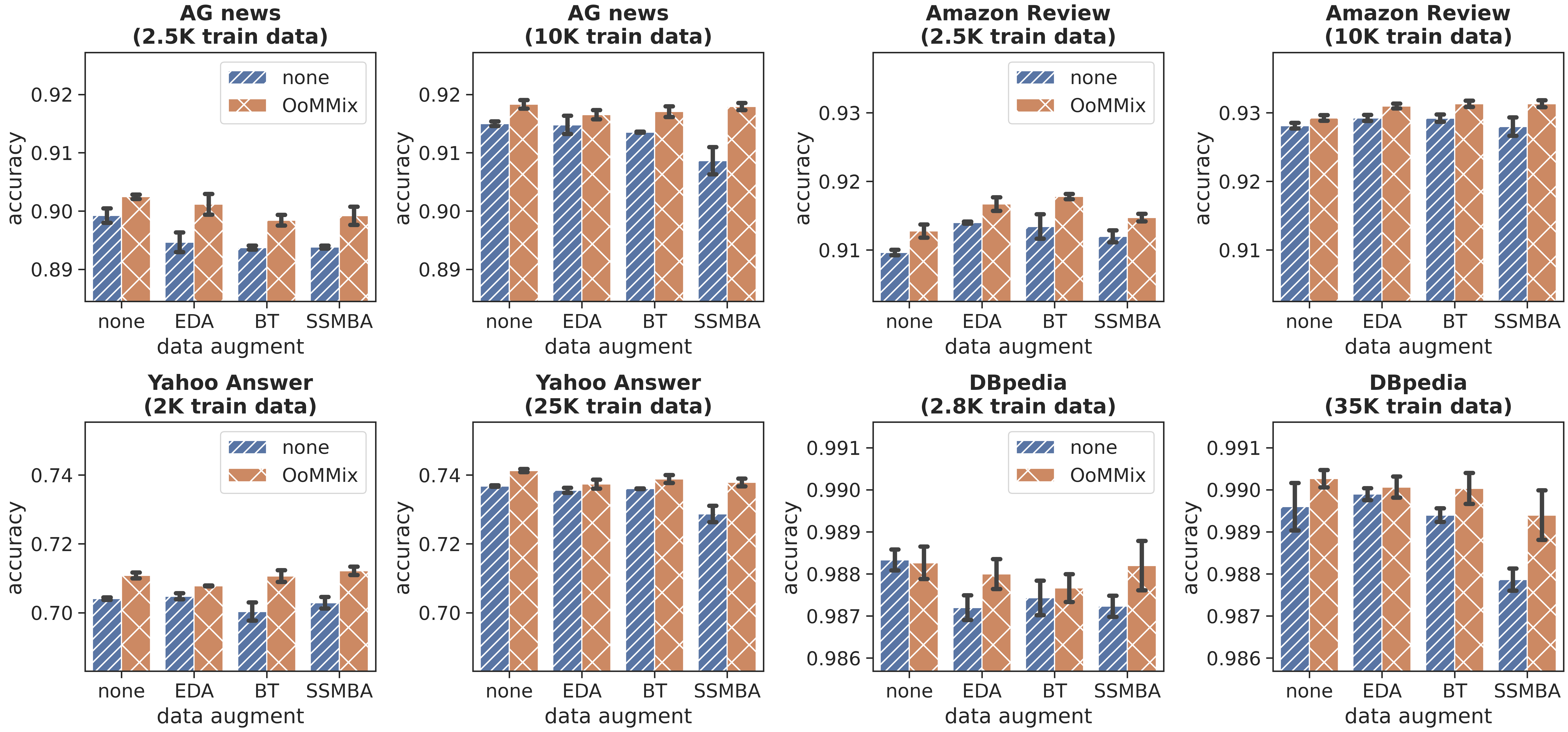}
    \caption{Average classification accuracy and their standard deviation when \proposed is applied with various data augmentation techniques.}
    \label{fig:compatibility}
\end{figure*}
\subsection{Comparision with Mixup Approaches}
We compare \proposed with existing mixup techniques. 
All the existing methods manually set the mixing coefficient, whereas we parameterize the linear interpolation by the embedding generator, optimized to produce out-of-manifold embeddings.
\begin{itemize}
    \setlength\itemsep{-0.3em}
    \item \textbf{\nonlinearmix}~\citep{guo2020nonlinear} samples mixing coefficients for each word from the beta distribution, while using neural networks to produce the mixing coefficient for the label. We apply this approach to BERT.
    
    \item \textbf{\mixuptransformer}~\citep{sun-etal-2020-mixup} linearly interpolates the sentence-level embedding with a fixed mixing coefficient. 
    The mixing coefficient is 0.5 as the paper suggested.
    
    \item \textbf{\tmix}~\citep{chen-etal-2020-mixtext} performs linear interpolation on the word-level contextual embedding space and samples a mixing coefficient from the beta distribution. 
    We select the best accuracy among different alpha configurations $\{0.05, 0.1\}$ for the beta distribution. 
    
    \item \textbf{\mixtext}~\citep{chen-etal-2020-mixtext} additionally utilizes unlabeled data by combining \tmix with its pseudo-labeling technique.
\end{itemize}

Table \ref{tab:overall} reports the accuracy on various sentence classification benchmarks.
In most cases, \proposed achieves the best performance among all the competing mixup approaches.
In the case of \nonlinearmix, it sometimes shows worse performance than the baseline (i.e., fine-tuning only on original data), because its mixup strategy introduces a large degree of freedom in the search space, which loses useful semantic encoded in the pre-trained weight.
The state-of-the-art mixup approaches, \tmix and \mixuptransformer, slightly improves the accuracy over the baseline, while showing the effectiveness of the mixup approach.
Finally, \proposed beats all the previous mixup approaches, which strongly indicates that the embeddings mixed by the generator are more effective for regularization, compared to the embeddings manually mixed by the existing approaches.
It is worth noting that \proposed obtains a comparable performance to \mixtext, even without utilizing additional unlabeled data.
In conclusion, discovering the out-of-manifold and applying mixup for such subspace are beneficial in contextual embedding space.

\subsection{Compatibility with Data Augmentations}
To demonstrate that the regularization effect of \proposed does not conflict with that of existing data augmentation techniques, we investigate the performance of BERT that adopts both \proposed and other data augmentations together.
Using three popular data augmentation approaches in the NLP community, we replicate the dataset as large as the original one to use them for fine-tuning.
\begin{itemize}
    \setlength\itemsep{-0.3em}
    \item \textbf{EDA}~\citep{wei-zou-2019-eda} is a simple augmentation approach that randomly inserts/deletes words or swaps two words in a sentence. 
    We used the official codes\footnote{\url{https://github.com/jasonwei20/eda\_nlp}} with the default insertion/deletion/swap ratio the author provided.
    
    \item \textbf{BT}~\citep{xie2019unsupervised} uses the back-translation for data augmentation. 
    A sentence is translated into another language, then translated back into the original one. 
    We use the code implemented in the MixText repository\footnote{\url{https://github.com/GT-SALT/MixText}} with the checkpoint fairseq provided.\footnote{\texttt{transformer.wmt19.\{en-ru,ru-en\}.single} \texttt{\_model} are provided through the official torch hub.}
    
    \item \textbf{SSMBA}~\citep{ng-etal-2020-ssmba} makes use of the pre-trained masked language model.
    They mask the original sentence and reconstruct it by filling in the masked portion. 
    We use the codes provided by the authors\footnote{\url{https://github.com/nng555/ssmba}} with default masked proportion and the pre-trained weight.
\end{itemize}

Figure~\ref{fig:compatibility} shows the effectiveness of \proposed when being used with the data augmentation techniques.
For all the cases, \proposed shows consistent improvement.
Especially for the Amazon Review dataset, the data augmentation and our mixup strategy independently bring the improvement of the accuracy, because the subspaces targeted by the data augmentation and \proposed do not overlap with each other.
That is, \proposed finds out out-of-manifold embedding, which cannot be generated from the actual sentences, whereas the data augmentations (i.e., EDA, BT, and SSMBA) focus on augmenting the sentences whose embeddings are located inside the manifold.
Therefore, jointly applying the two techniques allows to tightly regularize the contextual embedding space, including both in-manifold and out-of-manifold.

\begin{figure}
    \centering
    \includegraphics[width=\linewidth]{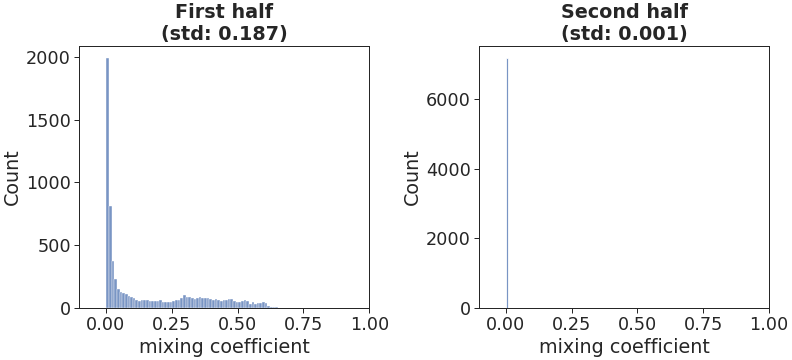}
    \includegraphics[width=\linewidth]{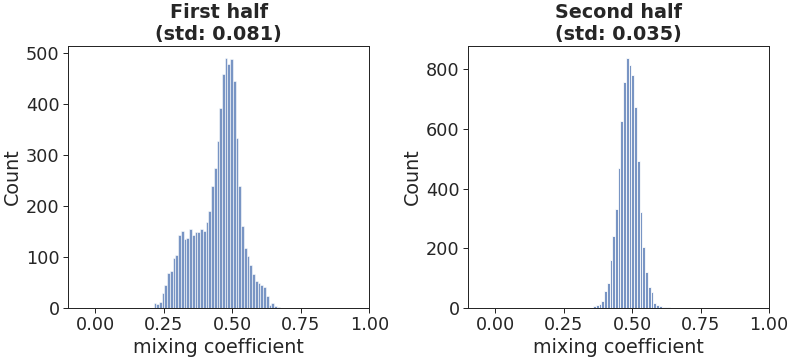}
\caption{Count of mixing coefficients without the discriminator (Upper) and with the discriminator (Lower).}
\label{fig:hist}
\end{figure}

Moreover, \proposed has additional advantages over the data augmentations. 
First, \proposed is still effective in the case that large training data are available.
The data augmentation techniques result in less performance gain as the size of training data becomes larger, because there is less room for enhancing the manifold constructed by enough training data.
Second, the class label of the augmented sentences given by the data augmentation techniques (i.e., the same label with the original sentences) can be noisy for sentence classification, compared to the label of out-of-manifold embeddings generated by \proposed.
This is because the assumption that the augmented sentences have the same label with their original sentences is not always valid.
On the contrary, there do not exist actual (or ground truth) labels for out-of-manifold embeddings, as they do not correspond to actual sentences;
this allows our mixup label to be less noisy for text classification.


\begin{figure}
    \centering
    \includegraphics[width=\linewidth]{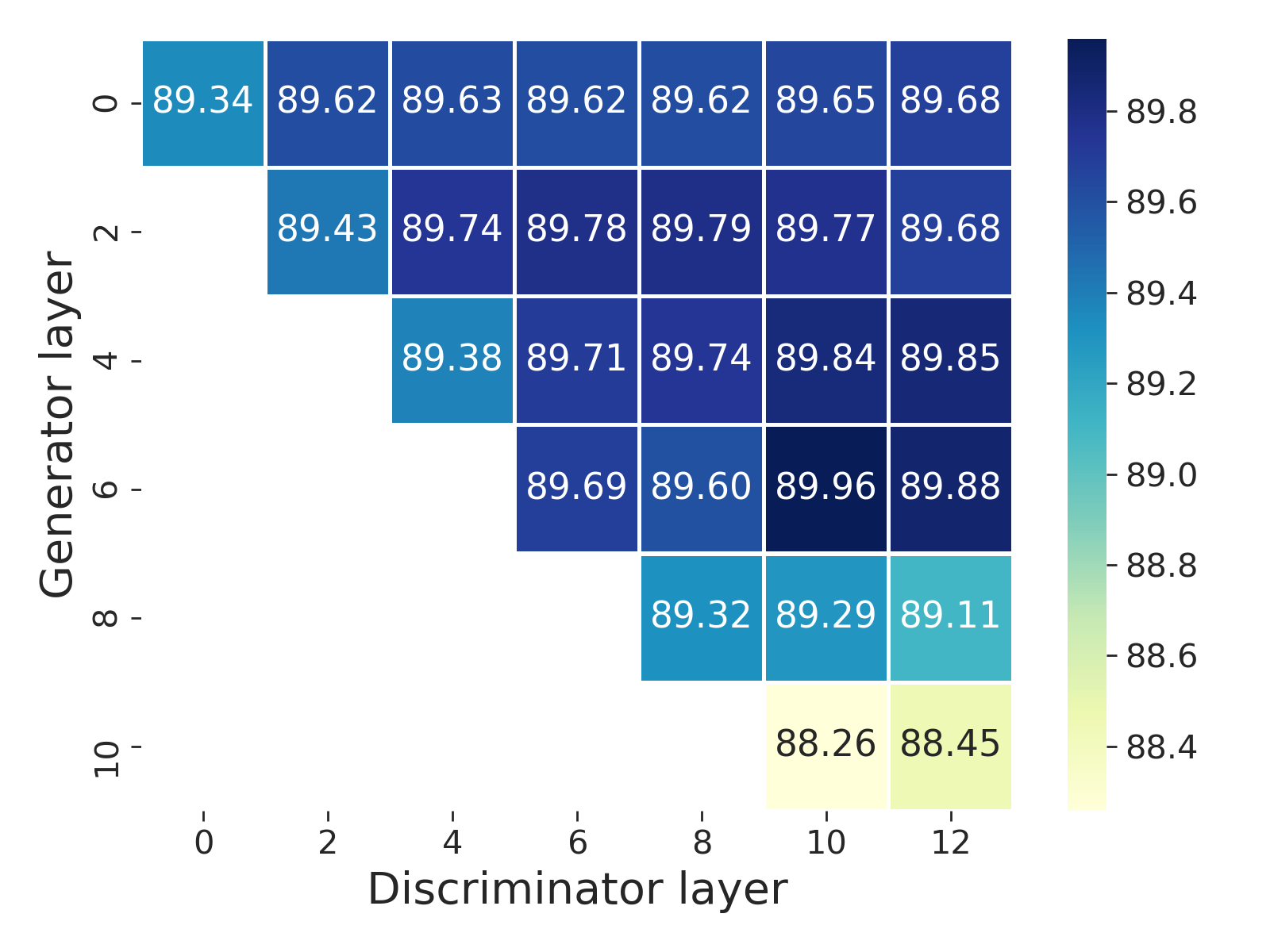}
    \caption{Performance changes with respect to different layers for the generator and discriminator.
    Dataset: Amazon Review 0.5K, Layer 0: word embedding.}
    \label{fig:sensitivity}
\end{figure}
\subsection{Effect of the Manifold Discriminator}
We also investigate how the manifold discriminator affects the training of the embedding generator.
Precisely, we compare the distributions of mixing coefficients, obtained from two different generators; 
they are optimized with/without the manifold discriminator, respectively (Figure~\ref{fig:hist} Upper/Lower).
We partition the training process into two phases (i.e., the first and second half), and plot a histogram of the mixing coefficients in each phase.

\begin{figure*}
    \centering
    \includegraphics[width=\linewidth]{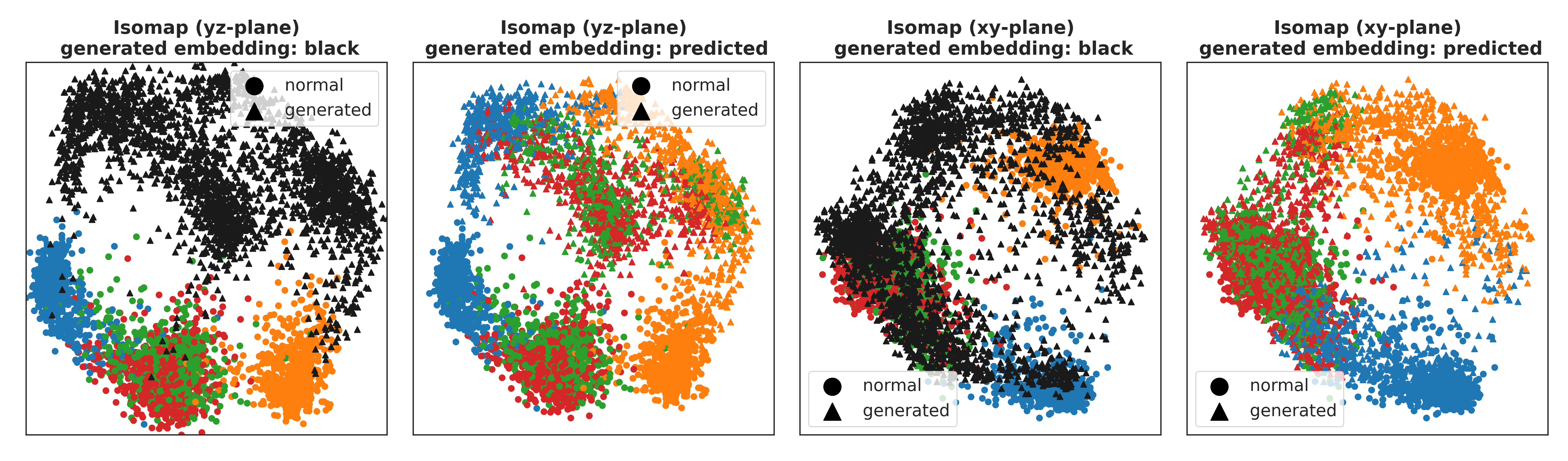}
    \caption{Isomap visualization of the sentence-level embeddings. The embedding vectors are projected into 3-dimensional space and rendered in two different views (xy, and yz-plane). For each view, we colorize the out-of-manifold embeddings with black and their predicted class.}
    \label{fig:vis}
\end{figure*}
The embedding generator without the discriminator gradually moves the distribution of the mixing coefficients toward zero, which means that the generated embedding becomes similar to the actual embedding.
Therefore, training the generator without the discriminator fails to produce novel embeddings, which cannot be seen in the original data.
In contrast, in the case of the generator with the discriminator, most of the mixing coefficients are located around 0.5, which implies that the generator produces the embeddings which are far from both the two actual embeddings to some extent.
We also observe that the average objective value for our discrimination task (Equation~\eqref{eq:oom_obj}) is 0.208 for the last 20 mini-batches;  this is much lower than 0.693 at the initial point.
It indicates that the generated embeddings are quite clearly distinguished from the ones computed from actual sentences.

\subsection{Effect of Different Embedding Layers}
We further examine the effect of the location of our generator and discriminator (i.e., $m_g$ and $m_d$) on the final classification performance.
Figure~\ref{fig:sensitivity} illustrates the changes of the classification accuracy with respect to the target contextual embedding layers the modules are attached to.
To sum up, BERT achieves high accuracy when the generator is attached to the contextual embedding lower than the sixth layer while the discriminator works for a higher layer.
It makes our out-of-manifold regularization affect more parameters in overall layers, which eventually leads to higher accuracy.
On the other hand, in case that we use both the generator and discriminator in the same layer, the gradient of the loss for manifold discrimination cannot guide the generator to output out-of-manifold embeddings,
and as a result, the generator is not able to generate useful embeddings.

\subsection{Manifold Visualization}
Finally, we visualize our contextual embedding space to qualitatively show that \proposed discovers and leverages the space outside the manifold for regularization.
We apply Isomap~\citep{tenenbaum2000global}, a neighborhood-based kernel PCA for dimensionality reduction, to both the actual sentence embeddings and generated embeddings.
We simply use the Isomap function provided by scikit-learn, and set the number of the neighbors to 15.
Figure \ref{fig:vis} shows the yz-plane and xy-plane of our embedding space, whose dimensionality is reduced to 3 (i.e., x, y, and z).
We use different colors to represent the class of the actual embeddings as well as the predicted class of the generated embeddings.

In the yz-plane, the actual sentence embeddings form multiple clusters, optimized for the text classification task.
At the same time, the generated embeddings are located in the different region from the space enclosing most of the actual embeddings.
In the second plot, we colorize the generated embeddings with their predicted class.
The predicted class of out-of-manifold embeddings are well-aligned with that of the actual embeddings, which means that \proposed imposes the classification capability on the out-of-manifold region as well.
We change the camera view to xy-plane and repeat the same process to show the alignment of class distribution clearly (in the third/fourth plots).
By imposing the classification capability on the extended dimension/subspace (i.e., out-of-manifold), \proposed significantly improves the classification performance for the original dimension/subspace (i.e., in-manifold).

\section{Conclusion}
\label{sec:conc}
This paper proposes \proposed to regularize out-of-manifold in the contextual embedding space.
Our main motivation is that the embeddings computed from the words only utilize a low-dimensional manifold while a high-dimensional space is available for the model capacity.
Therefore, \proposed discovers the embeddings that are useful for the target task but cannot be accessed through the words.
With the help of the manifold discriminator, the embedding generator successfully produces out-of-manifold embeddings with their labels.
We demonstrate the effectiveness of \proposed and its compatibility with the existing data augmentation techniques.

Our approach is a bit counter-intuitive in that the embeddings that cannot be accessed through the actual words are helpful for the target model.
As the discrete features from texts (i.e., words), embedded into the high-dimensional continuous space where their contexts are encoded, cannot cover the whole space, the uncovered space also should be carefully considered for any target tasks.
In this sense, we need to regularize the out-of-manifold to prevent anomalous behavior in that space, which is especially important for a large pre-trained contextual embedding space.



\section*{Acknowledgments}
This work was supported by the NRF grant funded by the MSIT (No. 2020R1A2B5B03097210), and the IITP grant funded by the MSIT (No. 2018-0-00584, 2019-0-01906).

\bibliographystyle{acl_natbib}
\bibliography{anthology,acl2021}


\end{document}